\documentclass{article}

\usepackage{PRIMEarxiv}

\usepackage[utf8]{inputenc} 
\usepackage[T1]{fontenc}    
\usepackage{hyperref}       
\usepackage{url}            
\usepackage{booktabs}       
\usepackage{amsfonts}       
\usepackage{nicefrac}       
\usepackage{microtype}      
\usepackage{lipsum}
\usepackage{graphicx}
\usepackage{subcaption}
\usepackage{caption}
\usepackage{multirow}
\usepackage{amsmath}
\usepackage{algorithm}
\usepackage{wrapfig}
\usepackage{algorithmic}
\usepackage{xcolor}
\graphicspath{{media/}}     
\makeatletter
\newcommand{\lora}{LoRA}
\makeatother

\makeatletter
\newcommand{\llama}{LLaMA}
\newcommand{\llamat}[1]{LLaMA 2-#1B}
\newcommand{\llamaa}[1]{LLaMA-#1B}
\makeatother

\newcommand{\falcono}{Falcon-rw-1.3B}
\newcommand{\spls}{30.1}
\newcommand{\splt}{20.1}
\newcommand{\splth}{10.1}
\newcommand{\opto}{OPT-1B3}
\newcommand{\stablelm}{StableLM-3B-4e1t}

\newcommand{\pl}{PrivateLoRA}
\newcommand{\bs}{bs}
\newcommand{\mcite}[1]{\cite{#1}}
\newcommand{\rpc}{r_{D2C}}
\newcommand{\rcp}{r_{C2D}}
\newcommand{\lmh}{LM Head}
\newcommand{\tpld}{T_{D}^{LRRT}}
\newcommand{\tplc}{T_{C}^{LRRT}}
\newcommand{\dcrun}{D\&C}
\newcommand{\tuni}{t}
\newcommand{\flopc}{F_{C}}
\newcommand{\flopd}{F_{D}}
\newcommand{\mbc}{MB_{C}}
\newcommand{\mbd}{MB_{D}}
\newcommand{\lrrt}{Low Rank Residual Transmission}

\newcommand{\nbcd}{B_{C2D}}
\newcommand{\nbdc}{B_{D2C}}

\newcommand{\tpspl}{TPS^{LRRT}_0}
\newcommand{\tpsdec}{TPS^{Decoder}_{C}}
\title{
  \pl~For Efficient Privacy Preserving LLM
  \thanks{Preprint. Work in progress.}
}

\author{
  Yiming Wang, Yu Lin, Xiaodong Zeng, Guannan Zhang \\
  Ant Group \\
  Shanghai, China\\
}

\begin{document}
\maketitle

\begin{abstract}

End users  face a choice between privacy and efficiency in current Large Language Model (LLM) service paradigms. In cloud-based paradigms, users are forced to compromise data locality for  generation quality and processing speed.
Conversely, edge device paradigms maintain data locality but fail to deliver satisfactory performance.
In this work, we propose a novel LLM service paradigm that distributes privacy-sensitive computation on edge devices and shared computation in the cloud. Only activations are transmitted between the central cloud and edge devices to ensure data locality. Our core innovation, PrivateLoRA, addresses the challenging  communication overhead by exploiting the low rank of residual activations, achieving over 95\% communication reduction. Consequently, PrivateLoRA effectively maintains data locality and is extremely resource efficient. Under standard 5G networks, PrivateLoRA achieves throughput over 300\% of device-only solutions for 7B models and over 80\% of an A100 GPU for 33B models. PrivateLoRA also provides tuning performance comparable to LoRA for advanced personalization. Our approach democratizes access to state-of-the-art generative AI for edge devices, paving the way for more tailored LLM experiences for the general public. To our knowledge, our proposed framework is the first efficient and privacy-preserving LLM solution in the literature\footnote{Demo and code is coming to github soon.}.
\end{abstract}


\section{Introduction}

In the rapidly evolving field of artificial intelligence, Large Language Models (LLMs) have emerged as a powerhouse\mcite{gpt3,llama,instructgpt,merge}.
Previously unsolvable long-tail problems are effectively tackled by LLMs, such as programming\mcite{codegeex,codellama}, instruction following\mcite{instructgpt,selfinstruct} and real world interaction\mcite{webgpt,palme,act1,act2}.
To fully harness the potential of large language models, it is crucial to concentrate on the privacy-preserving aspect of LLMs.
A critical dimension of this focus is data locality, which implies that user data are not stored on remote cloud servers but are instead kept closer to the user's own environment.
This approach not only reduces the risks associated with data breaches and unauthorized access but also aligns with growing global concerns over data sovereignty and user privacy.

However, like all other privacy preserving efforts, practicality of data locality is severely challenged by efficiency in the context of LLMs.
For current cloud-only and device-only LLM service solutions, end users are forced to choose between data locality and model scale.
Cloud-only centralized solution offers quick and quality generations from large scale models with its sufficient computing power.
But the data locality of personal data and personalized parameters are compromised.
Alternative decentralized solutions\mcite{mlcllm} like deploying smaller quantized\mcite{gptq} LLMs ($\leq$ 7B) on edge devices offer superior privacy but at the cost of performance.
Firstly, smaller models often fall short in delivering the emergent abilities exhibited by their larger counterparts\mcite{quant,merge}.
Secondly, on edge devices like smartphones, pure device solutions offer markedly low throughput.
For instance, a 3-bit quantized 7B model on a flagship smartphone achieves only 58.1 and 8.1 tokens per second for prefill and decode, respectively.
The prefill throughout is only 0.5\% of single A100 80G GPU, let alone comparing to clusters of high-end GPUs on Cloud.
Other efforts such as Federated Learning and Split Learning also face the challenge in efficiency, and do not solve the problem of inference.

After comparing Cloud-based and Device-based solutions, we confront a fundamental question: Can their benefits be synergistically combined to simultaneously achieve data locality and model scale?
Our solution is a heterogeneous distributed system\mcite{splitlearning} that leverages edge devices' storage for private data and personalized parameters, while utilizes the Cloud for computational enhancement.
Model parameters are split across the cloud and edge devices and only unreadable activations and gradients are transmitted to meet the requirement of data locality.
The largest challenge to build such heterogeneous systems is to use much smaller connection bandwidth to transmit equal amount of activations in homogeneous setups\mcite{zero,fsdp}.
The central cloud and edge devices are presumably connected via Internet, the bandwidth of which is significantly smaller than dedicated connection found in homogeneous setup ($1Gbps$ vs $ 100Gbps$).
The low connection bandwidth being the largest challenge, other challenges involve balanced workload distribution to tolerate significant internal performance gap.
For example, A100 GPU and flagship chipsets from smartphones can have over 20 times FLOPS difference and 40 times memory bandwidth difference.
The hardware performance gap can result in reduced throughput caused by blocking in pipeline.

To tackle the above challenges, we propose \pl, a novel Parameter Efficient Fine-Tuning (PEFT)\mcite{peft,delta1,delta2} method that exploits low rank of residual activations for communication and workload distribution.
Given the fact that low rank transforms on residual activation are enough to adapt transformer models, we make the hypothesis that decomposing one integral low rank transform into three sequential transforms yields comparable adaptation performances.
Therefore, \pl~integrates three sequential low-rank matrices ($A, M, B$) for weight adaptation (in Figure \ref{fig-met-pl}).
Non-trainable $A$ and $B$, deployed on the Cloud, serve as an encoder-decoder duo that condenses residual activations to reduce communication overhead.
We term this practice to reduce communication   \lrrt.
Trainable $M$ on edge devices transforms residual activations to steer the transformer for personalized outputs.
With less than 0.1\% of total parameters on edge devices, FLOPS and memory requirement are also largely reduced, yielding closer workload to processing power ratio between the cloud and edge devices.

On the basis of \pl, we propose a data locality preserving paradigm for LLM on a heterogeneous distributed system built on central cloud and edge devices.
Raw data and personalized parameters $M$ are kept on edge devices throughout training and inference. Only unreadable activations and gradients are transmitted.
Thanks to \lrrt, the communication overhead of activations are reduced by over 95\% percent, yielding comparable throughput to cloud-only solutions.

As a result, while totally respecting data locality, \pl~effectively harnesses large-scale LLMs on the Cloud with limited resources on edge devices and provides good adaptation performance for personalization.
\pl~has been tested across various scales and benchmarks, including GSM8K, MMLU, BoolQ, and HellaSwag. Despite the untrainable $A, B$ and the triplet structure, \pl~maintains tuning performance on par with the original \lora.
Our throughput estimations reveal that \pl~surpasses device-only solutions in both inference and training.
Utilizing average consumer-accessible network bandwidth and smartphones, \pl~achieves 175.5 and 26.5 tokens per second on 7B model for generation prefill and decoding, respectively, which are both over 300\% of device-only solutions.
Additionally, as model scales increase, \pl's advantages become more pronounced, achieving over 74\% of the throughput of an A100 80G GPU with a 33B model backend.
Our work is also orthogonal to previous efforts on efficient transformer inference, thus can be employed together to further boost efficiency.

In summary, the contributions of our paper can be summarized as,
\begin{itemize}
    \item We propose \pl, a novel PEFT method that achieves communication reduction and balanced workload distribution in distributed scenario.
    \item On the basis of \pl, we propose a new LLM service paradigm that heterogeneously distributes LLMs to protect data locality.
    \item Extensive empirical experiments are carried out to present a comprehensive study on the tuning performance, integrity and scalability of \pl.
    \item Numerical estimations show that with \pl, edge devices can achieve a throughput over 3 times of device-only solutions and comparable to running on GPU.
\end{itemize}

\section{Related Work}

\subsection{PEFT}
PEFT methods lowers hardware requirement of model fine-tuning by significantly reducing trainable parameters and consequently optimizer states cached in VRAM.
By exploiting the local optimum of a pretrained model, a much smaller solution space brought by reduce trainable parameters helps PEFT methods achieve comparable tuning performance\mcite{delta1,delta2}.
PEFT can be classified into two categories: 1) reparameterization-based methods\mcite{bitfit, rome} that retrain a portion of the parameters and 2) addition-based methods that train additional parameters\mcite{lora,adalora,adapter}.
Recent works in PEFT focus on resource efficiency\mcite{adapter,ia3,lora,adalora}.
\lora\mcite{lora} fits incremental weights by decomposing them into low-rank matrices. (IA)$^3$ tunes hidden states with learned multipliers.

\subsection{Distributed Learning For Data Privacy}

Federated Learning (FL) and Split Learning\mcite{splitlearning} are proposed to tackle the problem of data privacy in a distributed manner.
FL allows multiple nodes to locally train a complete neural networks without explicit exchange of local data via specialized optimization algorithms and transmission protocols.
FL has been widely applied in various domain such as computer vision, text typing (Google's G-board) and intrusion detection\mcite{dong2022interpretable,dong_ispa}.
Split Learning splits model vertically among different nodes so that only activations and gradients are transmitted to protect data locality.
In the context of LLMs, efficiency of these methods are highly questionable as both compute and communication are astrological compared to conventional deep neural networks.

\subsection{Running LLMs on Edge Devices}
Efforts have also been made to run LLMs on edge devices.
llama.cpp\footnote{https://github.com/ggerganov/llama.cpp} ports \llama~model in C/C++ so that models can be efficiently executed on  limited hardwares such as laptops and Raspberry Pi.
MLC Chat\mcite{mlcllm} showcases a model compilation solution for deploying language models on diverse hardware backends and applications.
With MLC Chat, quantized 7B model can be run on smart phones.
As for training LLMs on edge device, PockEngine\mcite{pock} introduces sparse back-propagation, pruning the backward graph, and updating the model sparsely to save memory and reduce latency while maintaining model quality.
PockEngine has demonstrated the capability to fine-tune LLaMav2-7B on NVIDIA Jetson AGX Orin with significant speed and memory efficiency compared to standard TensorFlow.

\section{Method}

\subsection{Problem Formulations}

\subsubsection{Notations}
Two heterogeneous runtimes with huge performance gap are in the scope, namely the cloud ($C$) and the edge device ($D$).
Cloud has powerful hardware and a large-scale shared autoregressive transformer model $P_{\Phi}(y|x)$ that produces non-personalized outputs.
Edge device has limited hardware and stores private data $Z=\{(x_i,y_i)\}$, where both $x_i$ and $y_i$ are sequences of tokens.
For example, $Z$ could be rounds of chat and $x_i$ and $y_i$ denotes messages of senders and receivers.
Hardwares are parameterized with FLOPS $F$ and memory bandwidth $MB$.
Two runtimes are connected with Internet with asymmetric network bandwidths denoted as $\nbdc$ and $\nbcd$. And we note such hybrid runtime $\dcrun$.
We obviously have $\flopc \gg \flopd$ and $\mbc \gg \mbd$.
For a paired connection, $\nbdc$ and $\nbcd$ are usually bound by edge device's Internet Service Provider, which result in $\nbcd > \nbdc$ but $\nbcd$ and $\nbdc$ are in the same order of magnitude.

Model $P_{\Phi}(y|x)$, parameterized by $\Phi$, is composed of an embedding layer, $N$ transformer layers and an \lmh~for token prediction.
The dimension of hidden states $h$ is noted as $d$.
For simplicity of discussion, we assume $M$ is decoder-only transformer but obviously \pl~also works with encoder architecture.

\subsubsection{Objectives of \pl}
\label{sec-obj-pl}

\lrrt~aims to solve the impossible triangle of performance, parameter-based personalization and data locality becomes for centralized or decentralized paradigm.
We then give formal definition on the three objectives.

\textbf{Performance} indicates both task solving capability and processing speed.
Since cloud-based solution fails in data locality, the performance to beat is benchmarked from device-only solutions.
\pl~has to outperform device-only solutions from both aspects.
\pl~is capable of leveraging large scale model on Cloud, therefore guarantees to outperform smaller models on edge devices in task solving capabilities.

Task solving capability is measured with conventional benchmark scores.
Processing speed is measured by number of processed tokens per second, or tokens per second (TPS) for short.
Transformer throughput largely depends on the sequence length and batch size, therefore throughput of generation prefill, generation decoding and batched training will all be considered.

\textbf{Parameter-based Personalization} achieves personalization by tuning knowledge or preferences into the models.
For example, in order to make model mimic receiver's tone, we can tune additional parameters $\Delta \Phi$ with chat records $Z$ while freezing $\Phi$. The objective can be written as,
\begin{equation}
    \arg\max_{\Delta \Phi} \sum_{(x,y)\in Z} \sum^{|y|}_{t=1} \log (P_{\Phi+\Delta \Phi}(y_t|x,y_{< t}))
\end{equation}

Compared to In-Context Learning (ICL) based personalization, the major advantage of parameter-based personalization is the capability to learn unseen concepts.
ICL-based personalization also increases sequence length and consequently computation and communication overhead, which may be critical to throughput on edge devices.

\textbf{Data Locality} requires that persistent storage of both raw data $Z$ and its derivative is only allowed on edge devices.
Raw data $Z$ is used as input for generation and input-label pair for autoregressive training.
Derivatives of data, especially personalized parameters, are necessary in training and inference.
Between the two, persistent storage of personalized parameters is more critical for centralized cloud-based solutions.
From the perspective of Cloud, storage of raw data is no longer required as long as the personalized parameters are obtained.
For \pl, this restriction also poses challenges on processing speed as it forces communication between edge devices and the cloud.

\subsection{\lrrt}
\begin{wrapfigure}{r}{0.4\textwidth}
    \centering
    \includegraphics[width=.4\textwidth]{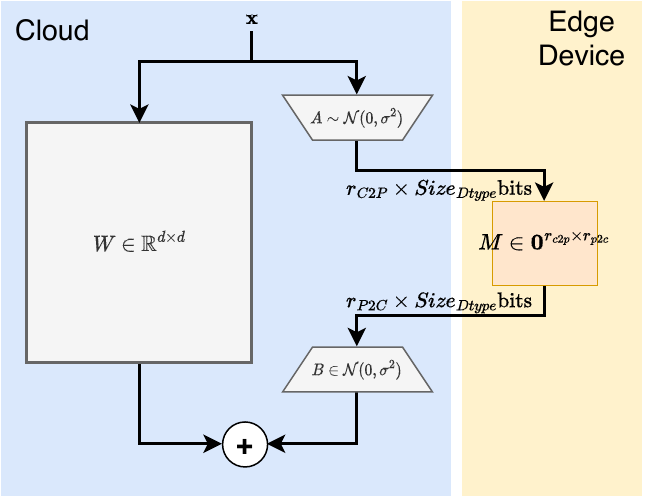}
    \caption{Weight adapted with \pl. Adaption matrices $A$ and $B$ are kept frozen after random initialization, thus contain no user information and can be stored and deployed on Cloud. Weight $M$ is trained with user data thus deployed on Phone. Thanks to \pl, the communication base multiplier is significantly smaller than model's hidden size and irrelevant to model architecture and scales. }
    \label{fig-met-pl}
\end{wrapfigure}

Given the fact that low rank transform is enough for effective LLM adaptation, \pl~exploits such low rank for communication and workload distribution.
Drawing inspiration from \lora, \pl~adapts model weights by adding three sequential low rank matrices parallel to target weight (see Figure \ref{fig-met-pl}).
Given  weight matrix $W\in \mathbb{R}^{d\times k}$ in target linear module and activations $\textbf{x}$ of dimension $d$, adaptation of \pl~can be written as,
\begin{equation}
    \textbf{x}W+\textbf{x} \Delta W=\textbf{x}W+ \textbf{x}AMB
\end{equation}
where $A\in \mathbb{R}^{d\times \rcp}$, $M\in \mathbb{R}^{\rcp\times \rpc}$ and $B\in \mathbb{R}^{\rpc \times d}$ . $\rcp,\rpc\ll d$, dimensions of $M$, are separately noted to account for the asymmetric bandwidth $\nbcd,\nbdc$.

Like all PEFT methods, the decoder stack of the model is frozen and only additional parameters are trainable.
In the adapted model forward, residual activations are firstly down-projected by $A$ to low rank for fast download, then transformed by $M$ while maintaining its low dimension for upload transmission, and finally up-projected by $B$ to merge residual activations into the base activation.
By dividing one integral low rank transform $\Delta W$ into three sequential transforms, we could explicitly exploit the low rank for communication.
Matrices $A,B$ serve as encoder-decoder pair to condense residual activations for transmission.

\subsubsection{Communication Overhead}
\pl~cuts down communication base from $d$ to $\rpc$ or $\rcp$ and make the transmission base number irrelevant to model scale.
Assuming the original \lora~is similarly deployed, transmission overhead of decoder stack would be proportional to the dimension of hidden states $d$.
$d$ is too big as transmission base and also scales up with model scale.
For example, $d$ equals 4096, 5120 and 6656 for \llamat{7}, \llamat{13} and \llamaa{30}, respectively.
Transmitting hidden states of one token of 16-bit precision on all decoder layers of \llamat{7} produces over 2Mb overhead, let alone multiple back-and-forth transmissions and much longer sequences in actual training and inference.
Whereas for \pl, assuming $\rcp=\rpc=128$, the base multiplier is cut down from 4096 to 128, a reduction of 96.88\%.
Another thing worth mentioning is that $\rcp,\rpc$ is invariant to model architecture, thus don't scale up with model scale.
The implication is that when leveraging larger models as personalization backend, the impact of communication overhead is more negligible.

\subsubsection{Workload Distribution}
In \pl, workload distribution is largely balanced so that a closer ratio of compute to processing power is achieved for Cloud and Device. On Cloud, apart from the original decoder stack, $A,B$ of \pl~adds very marginally additional compute. For \llamat{7}, assuming $\rcp=\rpc=64$, additional parameters take up 1.4\% of original model. Also, $A,B$ can be computed on parallel with $W$, thus no additional latency is introduced.
On edge devices, during forward computation of the decoder stack, the computation workload only involves matrix multiplication of small dimensions. For \llamat{7}, the parameter count of $M$ on edge devices take up less than 0.1\% of the original model, thus significantly reducing compute and memory pressure.
Moreover, the pure linearity of computation can yield higher utilization of the hardware.
Softmax operation in self attention is infamously slow due to low hardware utilization\mcite{flashattention,transinfer2} and reduces overall throughput in device-only solutions.
In our proposed \pl, only linear projections are performed on edge devices which are highly optimized from the perspective of both software and hardware.

\subsubsection{Data Locality}
During training, $A$ and $B$ are non-trainable and only $M$ is optimized with user private data.
$M$ is kept on Device during both training and inference.
Since $A$ and $B$ are randomly initialized and kept random, they contain no user data can also be stored on Cloud. Therefore, \pl~meets the requirement that no private data or its derivative are kept on Cloud.
Moreover, $A,B$ and $B$ form a tight pairing similar to public-private key used in encryption.
Possessing either component do not result in correct model output.

\subsubsection{Target Modules of \pl}
\label{sec-met-modules}
Several researches have pointed out that a thorough adaptation to every linear projection in transformers, including self attention and MLP, yields better overall performances\mcite{qlora}
We only target query, key and value projections in self-attention for adaptation to achieve minimized communication.
Target modules don't affect time complexity in single forward scenarios such as generation prefill and training, but can affect generation decoding.
During iterative sampling in generation decoding, query projection is only calculated with the newly generated token. Key and Value projections can reutilize KV cache to make marginal cost constant.
Whereas computation complexity of other linear projections are proportional to sequence lengths including both prompt and newly generated tokens.
Adapting these modules make decoding throughout decay due to increasing communication overhead.
Therefore, only adapting query, key and value modules produces consistent communication overhead in all scenarios.

\subsection{Paradigm Shift Powered by \pl}

\begin{figure}[tbp]
    \centering

    \begin{subfigure}[b]{.42\textwidth}
        \includegraphics[height=4.5cm]{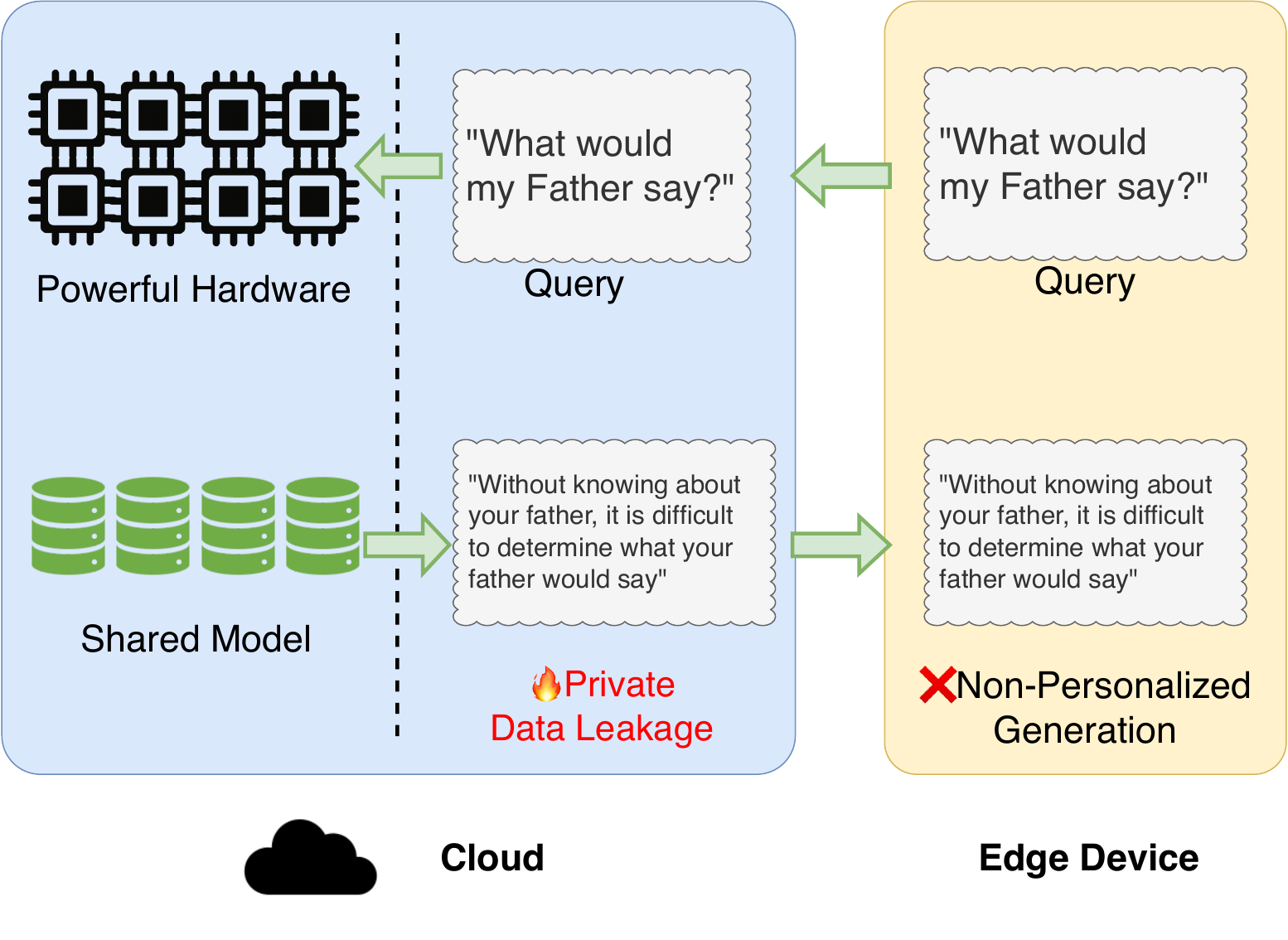}
        \caption{Centralized cloud-based paradigm}
        \label{fig-mode-cur}
    \end{subfigure}
    \begin{subfigure}[b]{.55\textwidth}
        \includegraphics[height=4.5cm]{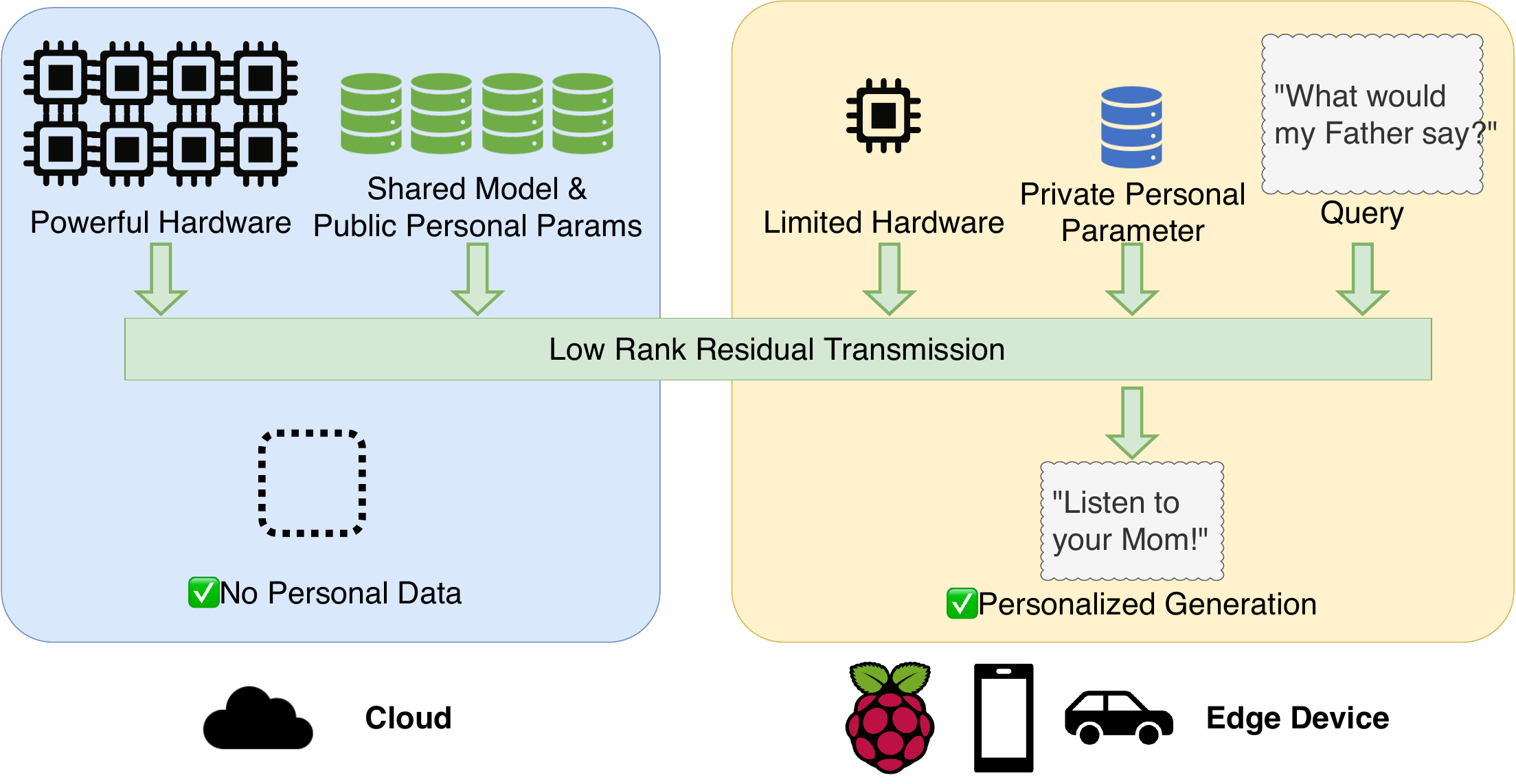}
        \caption{\pl-powered paradigm}
    \end{subfigure}
    \caption{Inference pipeline comparison between (a) centralized Cloud-based paradigm and (b) \pl-powered paradigm. \pl~has the advantages of data locality and personalized generation compared to Cloud-based paradigm.
        \pl~leverages powerful Cloud hardware to achieve parameter-based personalization with limited local resource on edge devices. Parameters containing personal data are stored on edge devices and remain on device during training and inference.}
    \label{fig-mode-versus-infer}
\end{figure}

Based on \pl, we propose new paradigm that solves the impossible triangle for efficient  LLM personalization.
\textbf{Parameter-based Personalization} Private personal parameters $M$ introduced by \pl~is optimized with personal data $Z$. Despite random and static $A,B$, residual activations can be effectively transformed by $M$ to output personalized generations.

\textbf{Model Performance} \pl~leverages models in cloud-based solutions as backend to guarantee better task solving capability compared to device-only solutions. Compared to cloud-based solutions, \pl~excels in providing tailored generations to better suit user demands.
In terms of throughput, thanks to \pl, communication overhead is significantly reduced.
Leveraging large scale LLM with \pl~can yield higher overall throughput compared to small models with device-only solutions.

\textbf{Data Locality}
Private personal parameters $M$, derivatives of raw data $Z$, are optimized with personal data but kept on Device in both training and inference.
Public personal parameters $A$ and $B$ are randomly initialized and kept frozen, thus not treated as derivatives of raw data.
For the locality of raw data, we deploy word embedding and \lmh~on Device and only the decoder stack on Cloud (see Figure \ref{fig-decode-cycle-pl}) so that only human-unreadable activations are transmitted (more discussion in Section \ref{sec-dis-act-sec}).
Therefore, data locality of both raw data and its derivative are protected with \pl.

\subsubsection{Inference and Training Cycle Powered by \pl}
\begin{wrapfigure}{r}{0.45\textwidth}
    \centering
    \includegraphics[width=.45\textwidth]{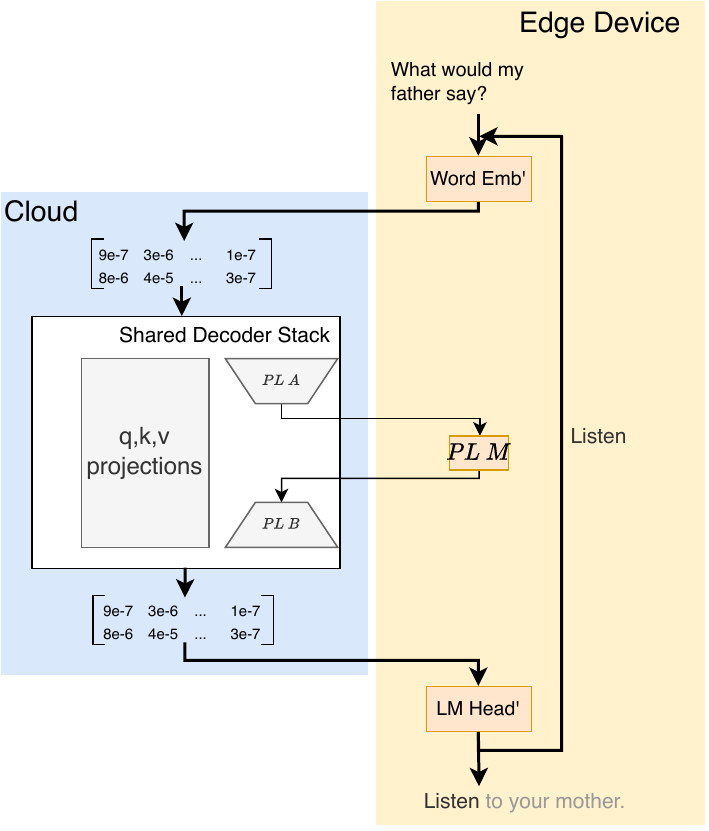}
    \caption{Decode cycle of decoder-only LLM generation with \pl. Raw texts are kept on device and only unreadable activations are transmitted. Random and static matrices $A,B$ on Cloud serve as activation encoder-decoder to cut down communication overhead. $M$ on Device steers the residual activations to produce personalized generation.}
    \label{fig-decode-cycle-pl}
\end{wrapfigure}
To fully understand the paradigm powered by \pl, we elaborate on the training and inference cycle of a heterogeneously distributed.

\textbf{Inference}
LLMs predict new tokens iteratively. Figure \ref{fig-decode-cycle-pl} plots the iterative sampling with \pl.
Starting with a query input by end user, the query is firstly tokenized and go through word embedding to get initial token embeddings.
These token embeddings are usable to central decoder stack but cannot be mapped back to token ids with embeddings on Cloud.
Token embeddings then go through series of decoder layers. In query, key and value projections in each self attention module, \pl~is applied to get residual activations from Device that ultimately produce personalized generation.
At the end of the decoder stack, activations of the last token are transmitted to Device and decoded by \lmh~on Device.
The newly obtained token are then tokenized and fed through word embedding for another round of sampling.

\textbf{Training}
In training cycles, $M$ is tuned with private user data for parameter-based personalization.
$A,B$ and the decoder stack is frozen.
Training cycle starts with forward computation almost identical to inference except that activations of the entire of the last decoder layer are transmitted to Device.
On Device, loss is calculated after feeding received activations to local \lmh.
After that, the loss is back propagated through the entire decoder stack.
Therefore, throughout training, personal data is always kept on Device and invisible to Cloud.

\subsubsection{Memory and FLOPs Analysis}
\label{sec-met-resource-ana}
\pl~is extremely friendly to edge devices with limited computation resources.
We numerically estimates required resource and compute to compare deploying complete transformer model and \pl~counterpart.
For memory requirement, we calculate the memory necessary on edge devices to load model parameters into RAM.
Total FLOPs on edge devices to complete one forward computation on one token are estimated.
The FLOPs are roughly estimated as 2 times of parameter count. Various datatype precisions are considered. And the results are listed in Table \ref{tab-resource-anal}.
With \pl, leveraging 7B models of 16-bit datatype only needs {10.6\%} of memory and {2.0\%} of FLOPs of 3-bit device-only solutions.
Furthermore, \pl's efficiency allows it to leverage models up to 30B in size with less memory and total FLOPs than 1B quantized models, underscoring its high resource efficiency.

\subsubsection{Throughput Analysis}
\label{sec-meth-infer-est}
In this section, we give detailed analysis on throughput in various scenarios.
We will decompose the elapsed time of forward computation into several well-known quantity in an end-to-end manner, so that the overall throughput of our proposed heterogeneous distributed system can be numerically estimated in Section \ref{sec-exp-infer-speed}.

We firstly start with forward computation and then extend our results to training.
For our proposed architecture, time of one forward computation $T$ can be decomposed into time of decoder stack on Cloud $T^{Decoder}_{C}$, time of \lmh~on Device $T^{\lmh}_{Device}$  and overhead introduced by \pl~$T^{\pl}$.
\begin{equation}
    \label{eq-og}
    T=T^{Decoder}_{C}+T^{\pl}+T^{\lmh}_{D}.
\end{equation}
Overhead of \pl~$T^{\pl}$ can be decomposed into network transmission $\tuni$ and local execution time on Cloud $\tplc$ and Device $\tpld$.
\begin{equation}
    \label{eq-pl-overhead}
    T^{\pl}=T^{\pl}_{Network} + \tpld + \tplc.
\end{equation}

Since network communication is more critical in our heterogeneous distributed system, we then focus on dissecting network overhead $T^{\pl}_{Network}$.
Assuming the number of adapted layer is noted as $N'\leq N$, the network communication can be decomposed into the initial and final embedding and decoder stack communication.
\begin{equation}
    \label{eq-2}
    T^{\pl}_{Network}=T^{TokenEmb}_{Network}+T^{\pl\ Activation}_{Network}\times N',
\end{equation}
where $T^{TokenEmb}_{Network}$ are transmission latency of input and output of decoder stack on Cloud, $T^{\pl\ Activation}_{Network}$ denotes transmission latency of \pl's activation.

Starting from Equation \ref*{eq-2}, we can then derive a unitary communication time $\tuni$ complete one forward computation on one token:
\begin{equation}
    \tuni= Size_{DType}\times (N'\times (\frac{\rpc\times N_{D2C}}{\nbdc} + \frac{\rcp\times N_{C2D}}{\nbcd}))+T^{TokenEmb}_{Network},
\end{equation}
where $N_{D2C}(N_{C2D})$ are number of D2C(C2D) transmissions per decoder layer and $Size_{DType}$ are number of bits of the used datatype.
$N_{D2C}(N_{C2D})$ are determined by variable dependency during forward computation.
For example, query, key and value projections share common inputs, thus $N_{C2D}$ is 1. They produce three activations of the same dimension, thus $N_{D2C}$ is 3.

\begin{wraptable}{r}{.45\textwidth}
    \centering
    \begin{tabular}{ll|rr}
        \hline
        \# Param             & \# Bit  & Memory(MB) & FLOPs (G) \\\hline
        \multirow{3}{*}{1B}  & 3       & 491.9      & 2.6       \\
                             & 4       & 655.8      & 2.6       \\
                             & 16      & 2623.3     & 2.6       \\\hline
        \multirow{2}{*}{3B}  & 3       & 999.0      & 5.3       \\
                             & 4       & 1333.2     & 5.3       \\
        \hline
        \multirow{3}{*}{7B}  & 3       & 2477.7     & 13.2      \\
                             & 4       & 3303.5     & 13.2      \\
                             & 16 (PL) & 265.3      & 0.27      \\
        \hline
        \multirow{3}{*}{13B} & 3       & 4819.4     & 25.7      \\
                             & 4       & 6425.8     & 25.7      \\
                             & 16 (PL) & 331.6      & 0.33      \\
        \hline
        \multirow{3}{*}{30B} & 3       & 12118.2    & 64.6      \\
                             & 4       & 16157.7    & 64.6      \\
                             & 16 (PL) & 431.9      & 0.43      \\
        \hline
    \end{tabular}
    \caption{Memory and FLOPs comparison between \pl~and full model deployment. Memory requirement are calculated to load all parameters and total FLOPs to compute a single token. \pl~ only needs \% of. FLOPs are calculated assuming batch size is 1 and sequence length is 1.}
    \label{tab-resource-anal}
\end{wraptable}

Therefore, the elapsed time in Equation \ref{eq-og} for single forward computation writes as,
\begin{equation}
    \label{eq-est-infer}
    T=T^{Decoder}_{C}+\tpld + \tplc+ T^{\lmh}_{D}+ \bs\times l\times\tuni ,
\end{equation}
where $\bs$ is the batch size and $l$ is the sequence length of mini-batch.

\textbf{Inference Throughput}
With Equation \ref{eq-est-infer}, we can compute the throughput measured in tokens per second.
Firstly, we note combined throughput for local executing $A,B$ and $M$ as $\tpspl$.
\begin{equation}
    \tpspl=\frac{\bs \times l}{\tpld + \tplc}.
\end{equation}

Based on Equation \ref{eq-est-infer}, the TPS of \pl~can be written as
\begin{equation}
    \label{eq-infer-tps}
    TPS=\frac{1}{\frac{1}{\tpsdec}+\frac{1}{\tpspl}+\frac{1}{TPS^{\lmh}_{D}}+\tuni},
\end{equation}
where $\tpsdec,TPS^{\lmh}_{D}$ indicates the throughput of decoder stack on cloud and \lmh~on Device, respectively.
Equation \ref{eq-infer-tps} holds for both generation prefill and generation decoding, so we can effectively estimate the inference throughput of generation prefill, single query decoding and batched query decoding.

\textbf{Training Throughput}
Training is much more complicated and many aspects can vary the time cost, such as the  optimizer, implementation of computation graph.
Following analysis in inference, we decompose the time to complete one training step in an end-end manner.
During back-propagation, gradients of the same dimension as activations are transmitted between Cloud and Device in reversed direction.
\begin{equation}
    \label{eq-comp-train}
    T=T^{Decoder}_{C}+\tpld + \tplc+ T^{\lmh}_{D}+ \bs\times l\times(\tuni+\tuni'),
\end{equation}
where $\tuni'$ represents transmission time of gradients which is different from $\tuni$ due to asymmetric network bandwidth and other terms include both forward and backward computations.
The TPS of \pl~during training is then,
\begin{equation}
    \label{eq-tps-train}
    TPS=\frac{1}{\frac{1}{\tpsdec}+\frac{1}{\tpspl}+\frac{1}{TPS^{\lmh}_{D}}+\tuni+\tuni'},
\end{equation}

Therefore, with Equation \ref{eq-infer-tps} and \ref{eq-tps-train}, we can derive the performance boundary between \pl~and pure device solutions in both generation and training scenarios.

\section{Experiments}
This part essentially answers the following two questions (1) Whether \pl~provides good tuning performances with  randomly initialized and non-trainable $A$ and $B$ (2) how fast is \pl~compared to pure device solution or even running on GPUs?
Other properties of \pl, such as scalability and integrity, are also studied to present a more comprehensive understanding.

\subsection{Experiment Setup}

\subsubsection{Benchmarks}
We use tuning performance on various benchmarks to prove that \pl~offers good data fitting capability for effective personalization.
Our benchmarks involve most commonly used benchmarks including natural language understanding, common sense reasoning and logic arithmetic.
For common sense reasoning, we use BoolQ\mcite{boolq}.
We evaluate involved methods with LM-Eval Harness\mcite{eval-harness}.
We report zero-shot accuracy except that 5-shot evaluation is adopted for MMLU.
Since we also value generation speed, we introduce a custom metric $M_{S}$ that awards high generation speed and performance improvement via tuning.
\begin{equation}
    M_{S}=\frac{TPS}{TPS_{C}}(M-M_{NT}),
\end{equation}
where $TPS$ represents generation speed of tested method, $TPS_{C}$ denotes generation speed on GPU, $M$ denotes average benchmark score of tested method and $M_{NT}$ denotes task performance of original model without tuning. Larger $M_{S}$ represents better overall performance.

\subsubsection{Generation Speed Benchmark and Estimation}
\label{sec-exp-infer-speed}
\begin{wraptable}{r}{0.4\textwidth}
    \centering
    \begin{tabular}{l|c}
        \hline
                                 & \textbf{Specifications} \\\hline
        \multirow{2}{*}{Device}  & $MB_{D}=42.7GBps$       \\
                                 & $FLOPS_{D}=15.8T$       \\\hline
        \multirow{2}{*}{Cloud}   & $MB_{C}=1935GBps$       \\
                                 & $FLOPS_{C}=312T$        \\\hline
        \multirow{2}{*}{Network} & $B_{d2c}=60Mbps$        \\
                                 & $B_{c2d}=100Mbps$       \\\hline
    \end{tabular}
    \caption{Assumptions for numerical estimation of inference latency of \pl. Edge Device specifications are equivalent to flagship chipsets of smart phones. Cloud specifications are at the level of single A100 80G GPU. Network bandwidths are at the level of 5G used by average consumers.}
    \label{tab-specs}
\end{wraptable}
We use decoding speed of single query generation, measured in tokens per second (TPS), as the throughput metric.
Throughput are measured on GPU and smart phones representing Cloud and Device runtimes, respectively.
Inference speed of \pl~is numerically estimated according to Equation \ref{eq-infer-tps} and modest assumptions on the hardware and network listed in Table \ref{tab-specs}. Edge Device specifications are equivalent to flagship chipsets of smart phones. GPU specifications are at the level of A100 80G GPU.
Network bandwidths are at the level of 5G accessible to average consumers\mcite{5gsurvey,5gsurvey2}.
Detailed calculation is discussed in Section \ref{sec-lat-esti}.

\subsubsection{Baselines}
Various device-only solutions are available. To demonstrate the extreme efficiency of \pl, we choose solutions that allow LLMs to run on smart phones as baselines, including 1) small models, 2) quantized models and 3) heterogeneously distributed \lora~on \dcrun.

\textbf{Small Models} We refer to models with parameters fewer than 3B as small models. Models under this scale can easily fit into smart phone's memory and can run even without parameter quantization. For this category, we include \opto\mcite{opt}, \falcono\mcite{falcon} and \stablelm\mcite{stable}.

\textbf{Quantized Models} Quantization technique is commonly used to run LLMs on low computation power devices. Quantization reduces parameter precision, thus significantly reduce memory workload and can utilize processor's low precision computation. For this category, we include 4-Bit quantized models of \falcono\mcite{falcon},\opto\mcite{opt}, \llamat{7} and \llamat{13}. We primarily use GPTQ\mcite{gptq} to quantize our model.
Aforementioned models except \llamat{13} can be tested on MLC Chat on mobile phones. Despite the availability on phones, we still include 4-Bit quantized \llamat{13} to offer a more comprehensive understanding of impacts of quantization on models.
Quantized models weights are either publicly downloadable resources or quantized with open-source software\footnote{Implemented with AutoGPTQ https://github.com/PanQiWei/AutoGPTQ}.

\textbf{\lora~on \dcrun} To better demonstrate the communication advantages of \pl, we adopt \lora~on \dcrun~runtime. Cloud only holds decoder stack parameters and Device only holds \lora~parameters. Activations are transmitted to complete forward computation but the communication base is dimension of hidden states. We include two \lora~configurations, one optimized for generation speed and another optimized for tuning performance.  \begin{itemize}
    \item \textbf{Speed Oriented} We apply the same latency analysis and get communication budget to make \lora~on \dcrun~reaches comparable speed to \pl. With obtained communication budget, we explore among several allocation strategies. We report score from the best performant configuration. We note this configuration as \textbf{\lora$^\text{S}$}. Details about the adaptation configurations and budget allocation are listed in Appendix \ref{sec-app-loras}.
    \item \textbf{Performance Oriented} The equivalent structure of adapting query, key and value projections in every decoder layer in \llama. We note this configuration as \textbf{\lora$^\text{P}$}
\end{itemize}

\subsubsection{\pl~Configurations}

We adapt only query, key and value projections found in every decoder layer of \llama~as mentioned in Section \ref{sec-met-modules}.
We set $\rpc=\rcp=128$ to achieve the balance between benchmark performance and generation speed.
Under our predefined conditions, \pl achieves \spls~tokens/s for \llamat{7}, \splt~token/s for \llamat{13} and \splth~token/s for \llamaa{30}.

\subsection{Experimental Results}
\label{sec-exp}
The following results are obtained with Bfloat16 if not mentioned.

\begin{table}[htbp]
    \centering
    \begin{tabular}{lll|c|ccccc|cc}
        \hline
        \textbf{Model}               & \textbf{Method}  & \textbf{Runtime} & \textbf{TPS} & \textbf{GSM8K} & \textbf{HSwag} & \textbf{BoolQ} & \textbf{PIQA} & \textbf{MMLU} & AVG. & $M_{S}$      \\\hline
        \multirow{2}{*}{\falcono}    & No-tuning        & Cloud            & 49.5         & 0.8            & 61.6           & 62.4           & 74.7          & 25.9          & 45.1 & 0.0          \\
                                     & 4Bit             & Device           & 20.1         & 0.0            & 45.6           & 59.5           & 74.8          & 24.6          & 40.9 & -1.7         \\\hline
        \multirow{2}{*}{\opto}       & No-tuning        & Cloud            & 49.5         & 1.0            & 54.0           & 59.6           & 69.0          & 25.0          & 41.7 & 0.0          \\
                                     & 4Bit             & Device           & 20.1         & 0.0            & 26.5           & 39.1           & 55.3          & 23.4          & 28.9 & -5.2         \\        \hline
        \multirow{1}{*}{StableLM-3B} & No-tuning        & Cloud            & 37.3         & 7.7            & 73.9           & 75.3           & 79.2          & 41.8          & 55.6 & 0.0          \\        \hline
        \multirow{5}{*}{\llamat{7}}  & No-tuning        & Cloud            & 37.2         & 14.6           & 75.9           & 77.7           & 77.8          & 45.3          & 58.3 & 0.0          \\
                                     & 4Bit             & Device           & 8.1          & 3.6            & 54.8           & 73.9           & 77.2          & 36.2          & 49.1 & -2.0         \\
                                     & \lora$^\text{S}$ & \dcrun           & 25.6         & 14.7           & 75.9*          & 81.7           & 77.8*         & 50.4          & 60.1 & 1.3          \\
                                     & \lora$^\text{P}$ & \dcrun           & 2.0          & 35.7           & 78.2           & 88.5           & 79.2          & 57.9          & 67.9 & 0.5          \\
                                     & PL (Ours)        & \dcrun           & 27.1         & 25.1           & 77.0           & 88.1           & 78.5          & 54.3          & 64.6 & \textbf{4.6} \\\hline
        \multirow{5}{*}{\llamat{13}} & No-tuning        & Cloud            & 27.8         & 23.5           & 79.4           & 80.5           & 79.1          & 54.8          & 63.5 & 0.0          \\
                                     & 4Bit             & Device           & -            & 6.1            & 78.6           & 80.8           & 78.2          & 51.4          & 59.0 & -            \\
                                     & \lora$^\text{S}$ & \dcrun           & 19.5         & 23.5*          & 79.4*          & 80.5*          & 79.1*         & 54.8*         & 63.5 & 0.0          \\
                                     & \lora$^\text{P}$ & \dcrun           & 1.3          & 42.7           & 80.1           & 89.7           & 79.7          & 61.4          & 70.7 & 0.3          \\
                                     & PL (Ours)        & \dcrun           & 20.5         & 36.5           & 80.0           & 88.9           & 79.3          & 58.4          & 68.6 & \textbf{3.8} \\\hline
        \multirow{4}{*}{\llama-30B}  & No-tuning        & Cloud            & 16.7         & 34.8           & 82.6           & 83.1           & 82.3          & 57.8          & 68.1 & 0.0          \\
                                     & \lora$^\text{S}$ & \dcrun           & 11.7         & 34.8*          & 82.6*          & 83.1*          & 82.3*         & 57.8*         & 68.1 & 0.0          \\
                                     & \lora$^\text{P}$ & \dcrun           & 0.7          & 51.9           & 82.8           & 90.5           & 82.6          & 63.3          & 74.2 & 0.3          \\
                                     & PL (Ours)        & \dcrun           & 12.9         & 46.7           & 83.3           & 88.4           & 82.6          & 59.8          & 72.2 & \textbf{3.1} \\\hline
    \end{tabular}
    \caption{\pl~offers good tuning performance and high throughput comparable to cloud-based solutions. Token generation speed denotes specifically throughput at decoding stage.
        Cloud speed is measured on single A100 80G GPU and Device speed is measured on flagship smart phones. \dcrun~speed is numerically estimated with modest assumptions listed in Table \ref{tab-specs}. Starred (*) scores indicate worse benchmark performance after tuning.
        $M_S$ is a custom metric based on the product of tuning improvement and generation speed. Large $M_S$ represents both good tuning performance and high generation throughput.
    }
    \label{tab-exp-plfastgood}
\end{table}

From Table \ref{tab-exp-plfastgood}, we can draw the following conclusions.

\textbf{\pl~outperforms device-only solutions in terms of both benchmark and processing speed.} Compared to small models used in device-only solutions, \pl~is capable of leveraging large scale model as the tuning backend, thus guarantees better task performances.
For example, with the target generation speed of around 20 tokens per second, device-only solutions can only leverage 1B model while \pl~is capable of leverage 7B model.
Also, \pl~can achieve higher generation throughput with backend of 33B model compared to 7B model of device-only solutions, with a huge gap of 23.1 in average benchmark scores, let alone smaller scale models.

\textbf{Significantly reduced communication overhead allows better adaptation for \pl.} Combining results of \lora$^\text{S}$ and \lora$^\text{P}$, we find extreme disparity in tuning performance and throughput. For \llamat{13}, \lora$^\text{S}$ fails to improve the benchmark score of every tested task while \lora$^\text{P}$ notably boosts task performance of GSM8K and MMLU. On the other hand, the throughput of \lora$^\text{P}$ is even lower than device-only solutions.
Such observed disparity indicates the importance of number of adapted layers. To achieve comparable generation speed, \lora$^\text{S}$ only adapts two layers due to large communication base number of $d$.
But in the case of \pl, thanks to significantly reduced  communication base, all layers can be adapted and yield good tuning performances.

\textbf{Tuning performance of \pl~is comparable to \lora, thus offering good foundation for personalization.} On tested tasks, we find very close average benchmark scores between \pl~and \lora$^\text{P}$.
The gap between \pl and \lora$^\text{P}$ are 3.3, 2.1 and 2.0 for \llamat{7}, 13B and 33B model, respectively. In the following ablation study of scaling up ranks, we find the performance gap is even smaller, indicating \pl~offers reliable tuning performance which can be good foundation for personalization on private data.

\subsection{Ablation Study: Integrity of \pl}
\label{sec-exp-integrity}
\begin{wrapfigure}{r}{0.5\textwidth}
    \centering
    \includegraphics[width=.5\textwidth]{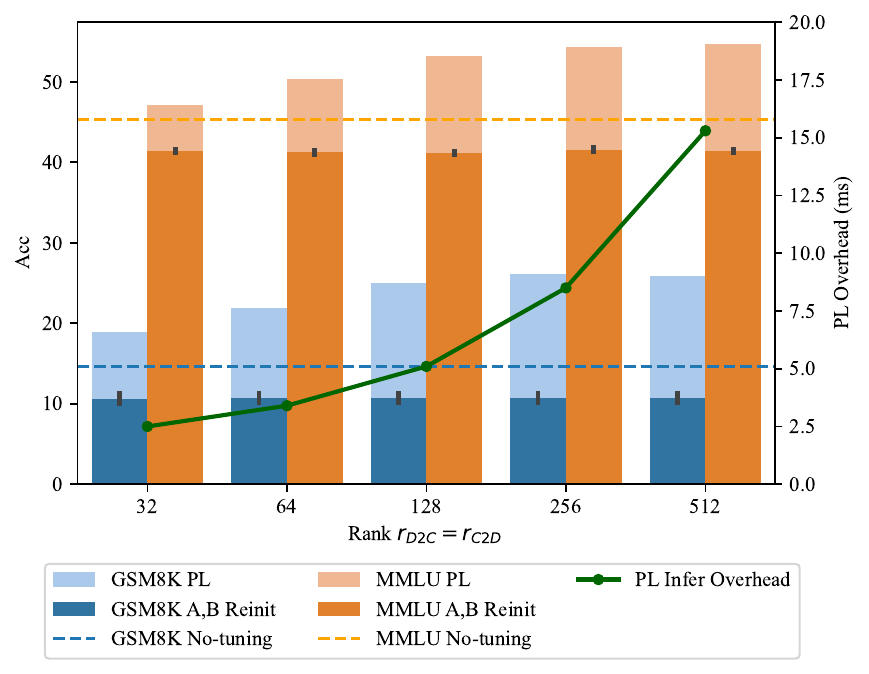}
    \caption{Benchmark scores, \pl~integrity and inference overhead of \pl~with scaled up ranks $\rpc,\rcp$. Unlike \lora, scaling up ranks benefits tuning performance. }
    \label{fig-exp-scalerank}
\end{wrapfigure}
$A,B$ are kept random after initialization and serve as activation encoder and decoder.
Despite the static nature of $A,B$, $A,B$ and $M$ form a matched integral pair during optimization of $M$.
We make the hypothesis about integrity of \pl~that with either of the pair, the model performance with degrade as the residual activations are transformed into noise.
In order to prove the integrity of $A,B$ and $M$ pair, we conduct the following ablation study.
For a tuned pair of $A,B$ and $M$, we reinitialize $A,B$ and keep $M$ intact. We then run the benchmark to see impact of our perturbation on performance.
We test on two benchmarks, namely MMLU and GSM8K, with \llamat{7}.
For each setup, we sample scores for 20 rounds.

Figure \ref{fig-exp-scalerank} plots the benchmark scores of \pl-tuned (lighter color), reinitialized $A,B$ (dark color) and original performance without tuning.
MMLU of reinitialized $A,B$ are around 40.8 and GSM8K around 10.7, which is inferior to original performance.
Overall, the poor performance after reinitializing $A,B$ verifies the integrity of $A,B$ and $M$.

\subsection{Ablation Study: Ranks of \pl}
We empirically choose $\rpc=\rcp=128$ in our main experiment.
To study the effect of ranks $\rpc,\rcp$, we train \llamat{7} on MMLU and GSM8K with gradually increased $\rcp,\rpc$ from 32 to 512. For simplicity, we set $\rcp=\rpc$.
The benchmark scores and resulting inference overhead are plotted in Figure \ref{fig-exp-scalerank}.

Unlike \lora, increasing ranks $\rpc,\rcp$ of \pl~increases tuning performances. When scaling ranks from 32 to 128, we see obvious improvement on benchmark scores of MMLU and GSM8K.
The benefit in tuning performance  decreases as ranks increase and we do not see much of a difference between ranks of 256 and ranks of 512.
With this in mind, we can effective adjust the tuning capability on demand.
Also, we found that with an equal rank of 32, \pl's benchmark score is way lower than \lora, indicating the impact of untrainable $A,B$ on data fitting capability.
However, with increased ranks, as the residual activation becomes of higher rank, the tuning performance is compensated.

\begin{wraptable}{r}{.4\textwidth}
    \centering
    \begin{tabular}{ll|cc}
        Method                            & Config                 & GSM8K         & MMLU          \\\hline
        \multirow{4}{*}{PL}               & $A\times$~~~$B \times$ & 25.1          & 54.3          \\
                                          & $A\times$~~~$B\surd$   & 26.7          & 54.5          \\
                                          & $A\surd$~~~$B\times$   & 27.4          & 55.4          \\
                                          & $A\surd$~~~$B\surd$    & \textbf{28.8} & \textbf{56.6} \\\hline
        \multirow{1}{*}{\lora$^\text{P}$} & q,k,v                  & 31.3          & 57.3          \\\hline
    \end{tabular}
    \caption{Ablation study on structure of \pl.
        Various $A,B$ configurations are tested.
        Trainable $A,B$ of \pl~benefits tuning performances.
        $A\times,B\surd$ means $A$ is non-trainable and $B$ is trainable.}
    \label{tab-exp-ab-trainab}
\end{wraptable}
Another thing to account for when setting the ranks of \pl~is the communication overhead. The inference overhead should increase linearly with ranks and Figure \ref{fig-exp-scalerank} verifies it.

\subsection{Ablation Study: Comparison with \lora}

Although \pl~is equally low rank as \lora, \pl~exhibits weaker tuning performance compared to \lora$^P$.
To find out whether the performance difference comes from triplet structure or untrainable $A,B$ , we then carry out an ablation study.

We run all four configurations of $A,B$ with different trainable setting for \pl. For \lora, we run the equivalent structure that only targets query, key and value projections. We benchmark on MMLU and GSM8K as \pl~falls behind compared to \lora.

Table \ref{tab-exp-ab-trainab} lists the benchmark scores of aforementioned methods.
We see largest improvements brought by simultaneously trainable $A,B$.
Another interesting thing is that trainable $A$ boosts overall performances more than trainable $B$, meaning a trainable upstream is more preferable.

\section{Throughput Estimation}
\label{sec-lat-esti}

According to TPS derived in Section \ref*{sec-meth-infer-est}, assumptions in Table \ref{tab-specs} and measured base throughput (Table \ref{tab-app-tpsm-train}), we can make numerical estimations on  throughput of \pl.
We discuss throughput of typical scenarios of transformer usage, including generation prefill, generation decoding and training.
To simplify the discussion and better present the idea, we use a single A100 80G GPU as Cloud.
Therefore, we do not include models larger than 33B.
Detailed calculation is discussed in Appendix \ref{sec-app-esti-tps}.

\subsection{Inference Throughput}

\begin{figure}[tbp]
    \centering

    \begin{subfigure}[b]{.48\textwidth}
        \includegraphics[width=\textwidth]{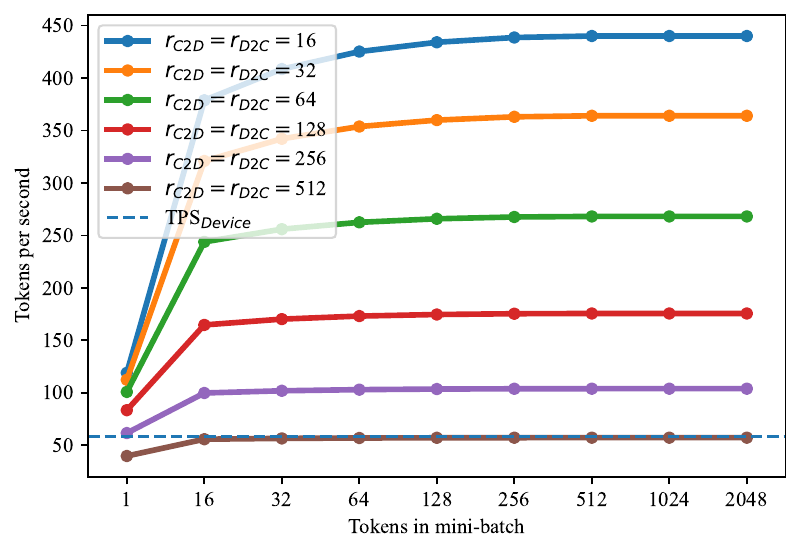}
        \caption{Prefill.}
    \end{subfigure}
    \begin{subfigure}[b]{.48\textwidth}
        \includegraphics[width=\textwidth]{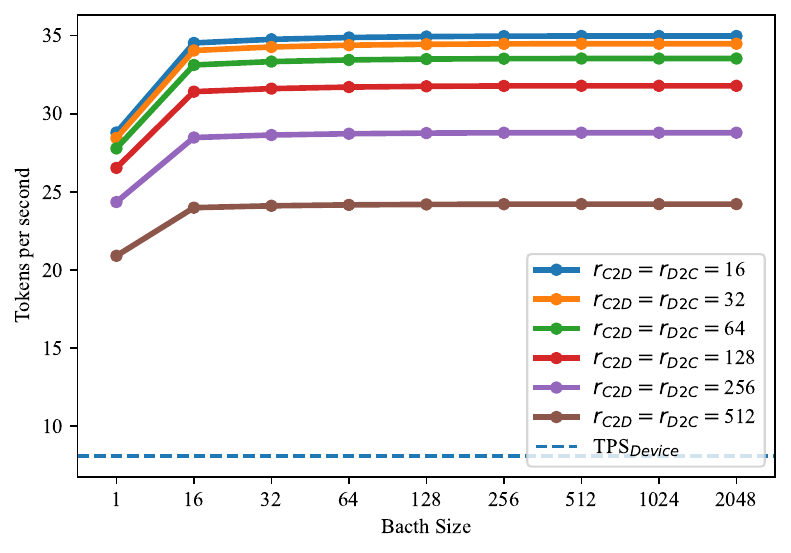}
        \caption{Decoding.}
    \end{subfigure}
    \caption{Inference throughput of \pl~with backend of \llamat{7}. Prefill and decoding speed  for rank of 128 reaches 175.5 and 26.5 tokens per second,respectively. Throughput of both stages is almost 300\% of pure device solutions.}
    \label{fig-lat-infer-speed}
\end{figure}

Figure \ref*{fig-lat-infer-speed} plots estimated inference throughput of prefill stage and decoding stage of \pl~with backend of \llamat{7} (more at Appendix \ref{sec-app-infer-tps}).
In general, TPS of \pl~increases with total tokens processed and converges when total tokens exceed certain thresholds.
With ranks set to 128, \pl~can achieve a prefill speed of 175.5 token per second, almost 300\% of pure device solutions.
As for decoding speed, \pl~allows a speed of 26.5 TPS, 327\% of pure device solutions.
If ranks are reduced to 64 or 32, the prefill TPS will be 268 or 364, respectively.
Therefore, the tradeoff between tuning performance and throughput is important in \pl~application.

\begin{wrapfigure}{r}{0.5\textwidth}
    \centering
    \includegraphics[width=.5\textwidth]{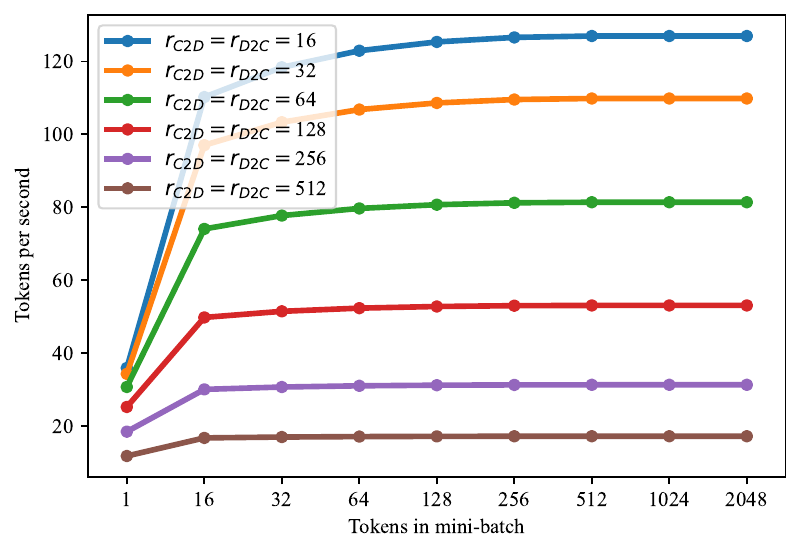}
    \caption{Estimated training throughput of tuning \llamat{7} with \pl~with different ranks and total tokens in mini-batch. Word embedding, adaptation matrices $M$ and \lmh~are set trainable. }
    \label{fig-exp-train-throughput}
\end{wrapfigure}
On larger scale models, execution time of the decoder layer is larger, therefore the impact of \pl~on throughput becomes smaller.
30B models are impossible to run on current smart phones.
However, it can be utilized as personalization backend of \pl, and throughput can reach 77.2\% of A100 80G GPU speed.
Although we do not offer numerical estimations on larger scale model, \textit{e.g.} 65B and 160B, due to the single A100 80G limit, we can still conclude that \pl~becomes even more competitive on models of these scales.

\subsection{Training Throughput}
\label{sec-dis-batch}
Training step is much more complex than pure inference. Compared to inference, model training not only adds back-propagation but also involves computations from optimizer, computation graph, \textit{etc}, thus adding highly variable workload to processor and memory.
All these aspects make estimation of time cost of training step difficult.
Moreover, since there are no publicly available deep learning frameworks that allow tuning LLMs on smart phones, we could not offer empirical training throughput on Device.
However, with our estimations, we still find that \pl~is highly competitive in training efficiency.

Estimated throughput of tuning \llamat{7} with \pl~of different ranks is plotted in Figure \ref{fig-exp-train-throughput}.
With rank set to 128, the peak training throughput of \pl~on \llamat{7} is around 53.8 tokens per second, which is almost as fast as prefill speed on Device.
On GPU, training throughput is way smaller than prefill throughput and the ratio sits around 10\% according to our benchmark.
Despite that we can not numerically estimate training throughput on pure Device, we can still conclude that tuning with \pl~outperforms pure device solutions in terms of speed.

\subsection{\pl~Overhead}
In this section, we analyze the overhead introduced by \pl, so that throughput under other conditions can be better predicted. In general, despite our modest assumptions, \pl~introduces very marginal overhead and can produce higher overall throughput if accompanied with SOTA LLM serving technology.

\begin{wrapfigure}{r}{0.5\textwidth}
    \centering
    \includegraphics[width=.5\textwidth]{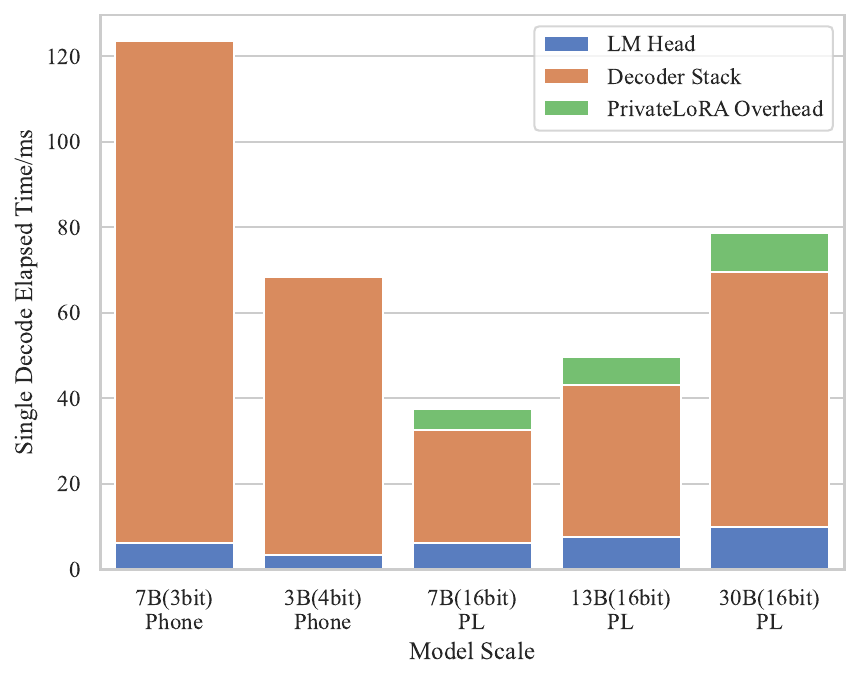}
    \caption{Inference latency decomposition shows that \pl~adds very marginal overhead. Under 5G accessible to average consumers, \pl~adds 5.1ms, 6.4ms and 9.3ms of overhead for 7B, 13B and 30B models, respectively. \pl~allows personalized generation with a backend of 30B models at the speed of 4bit quantized 3B models.}
    \label{fig-exp-infer-overhead}
\end{wrapfigure}

Figure \ref*{fig-exp-infer-overhead} plots detailed inference time decomposition.
Under 5G accessible to average consumers, \pl~adds 5.1ms, 6.4ms and 9.3ms of overhead for 7B, 13B and 30B models, respectively
Inference latency overhead introduced by \pl~is quite marginal and is only proportional to number of adapted layers.
Architecture of \pl~reduces transmission base from hidden dimension to ranks $\rpc,\rcp$ , so that transmission amount is model invariant.
Forwarding one token through one decoder layer only yield 8.2 Kb of transmission, while 262.1 Kb is needed in \lora~equivalent architecture and the amount is also relevant to hidden dimension.

Therefore, the marginal latency brought by \pl~allows for higher throughput if accompanied with dedicated LLM serving technology.
Throughput on GPU of this work is measured with Huggingface transformers\mcite{transformers} implementation, which is not dedicated for LLM serving.
SOTA LLM serving technology\mcite{vllm,mlcllm} can produce much higher throughput with enhanced FLOPS utilization and multi-GPU support. For instance, running \llamat{70} with MLC LLM\mcite{mlcllm} on 8 A100 80G GPU yields a decoding speed of 38.8 TPS.

\section{Discussion}
In previous sections, we discussed the architecture of \pl, its impact on LLM service mode and its sheer tuning performance.
In this section, we discuss other challenges of \pl~throughout the entire lifecycle and potential solutions.
Solutions are not limited at the level of models since many problems can be better handled in a more systematic manner.

\subsection{Further Improve \pl~Throughput}
Various methods can be adopted to improve \pl~throughput.

With current configuration, the transmission amount per token for \llamat{7} is 4.2Mb.
Following the overhead  decomposition in Section \ref{sec-meth-infer-est}, we can easily propose several approaches to further reduce the transmission overhead.
Firstly, a lower precision data type can be used. Various researches have pointed out that 8bit, even 4bit quantization does not influence much model performance and the impact reduces as model scales up. Therefore, it's reasonable to use \pl~on a lower precision model and the transmission overhead can be reduced.
Secondly, exploiting the redundancy in adaptation configurations. Adapting all modules at all layers may not be optimal and reduction in adapted modules can provide similar performances. Therefore, we can trade fractions of tuning performance with fewer adapted modules. With fewer adapted modules, the activation transmission is also cut down.
Thirdly, we can deploy some of the decoder layers on Device. For devices with reasonable hardware, deploy some decoder layers on Device can better utilize the hardware and reduce the transmission amount. As mentioned earlier, this practice also enhances the activation security.

\subsection{Activation Privacy}
\label{sec-dis-act-sec}
In our work, we protect privacy by ensuring data locality, which refers to the prohibition of persistent storage of raw data and its derivative on Cloud.
With \pl, raw text and personalized parameters remain on edge devices no matter in training or inference.
This protects the user from being identified by visual-based or norm-based comparison\mcite{DBLP:conf/icml/DongZL22}.
In our proposed paradigm, activations and gradients are transmitted to prevent raw data is uploaded to Cloud.
This contrasts with existing literature that focuses on extracting bag of words from general-purpose language model embeddings\mcite{privacy}.
However, such method targets final output embeddings and effect on intermediate activations of decoder-only generative LLMs is unknown.
To further enhance privacy, we can deploy critical decoder layers on edge devices to reduce the exposure of vulnerable activations\mcite{dola}.
Moreover, our proposed heterogeneous architecture design complicates potential attacks.
Looking ahead, our future work will concentrate on bolstering activation privacy, exploring more robust methods to protect against sophisticated data breaches.

\subsection{\pl~Integrity}

In Section \ref{sec-exp-integrity}, we confirm the integrity of the model components A, B, and M, showing that only matched pairs produce correct outputs. Two critical observations emerge from our analysis. First, we observed an intriguing phenomenon: the perturbed benchmark scores of GSM8K and MMLU both decline by approximately 5\% in accuracy compared to the original model, regardless of the noise activation rank. This finding indicates that increasing the rank of noise activations does not further degrade performance. Intriguingly, higher ranks improve tuning performance.
Starting from this observation, we may gain valuable insights into the mechanisms of low-rank adaptation of LLM.

Second, we identify a potential vulnerability (e.g., backdoor attack on edge\mcite{dong_infocom23}) in the model's design: the simple linear nature of components $A, B$, and $M$ could allow for 'hacking' of $M$, assuming $A$ and $B$ are known.
By tracking several rounds of forward computation, one might deduce $M$, leading to privacy concerns.
However, it's crucial to note that the parameters on Device are essentially a black box to the Cloud, and their internal architecture could be more complex than a single linear projection.
This realization directs our future work towards enhancing the security and integrity of these model components to achieve responsible\mcite{dong_ndss_2023}, secure and ethical use of LLM.

\subsection{Low power consumption is the core advantage of \pl.}

As detailed in Section \ref{sec-met-resource-ana}, \pl~necessitates a mere 2\% of the FLOPs required by a device-only solution, presenting a significant computational advantage.
This reduction in computation not only enhances processing speed but also leads to a substantial decrease in power consumption.
The key benefit of \pl~lies in its ability to minimize power usage on edge devices.
While advancements in hardware architecture and manufacturing processes may enable larger scale transformers to operate on edge devices, \pl's ability to drastically reduce local compute demands remains its standout feature.
This is particularly crucial for most edge devices, which aren't typically built for sustained peak performance and where power efficiency is paramount, especially in battery-dependent scenarios. Comparing \pl~with device-only solutions, there is a notable shift in power consumption from edge devices to the broader network and Cloud infrastructure.
Expanding on this, the adoption of \pl~could have profound implications for the future design and functionality of edge devices.

\section{Conclusion and Future Work}

In this paper, we propose \pl, a Parameter Efficient Fine-Tuning (PEFT) method for heterogeneously distributing LLMs.
This novel approach, centered around the concept of \lrrt, significantly diminishes communication overheads, thus offering a more efficient and privacy-conscious alternative to traditional cloud-based solutions.
Our proposed method democratizes access to advanced LLM capabilities and could spur a wave of innovation and new applications across various sectors.

The following directions can be our future work. \begin{itemize}
    \item \textbf{Communication Budget Allocation} In our experiment, we adapt all decoder layers, but redundancy in adaptation can be exploited to further reduce communication overhead.
    \item \textbf{Activation Privacy} Activation privacy is the next challenge towards a more comprehensive privacy protection in LLM service.
    \item \textbf{Integrity of $A,B,M$} Unmatched $A,B$ and $M$ only slightly reduces performance but it should work like public private key pair that unmatched pair results in block of access. A method should be found to completely poison the activations so that even semantics can not be deduced.
\end{itemize}



\bibliographystyle{unsrt}
\bibliography{references}

\newpage
\appendix
\section{Experiment Setup}
\subsection{Hyperparameters}
\label{app_hparam}
Detailed hyperparameters are listed in Table \ref{tab-app-hparam}.
All experiments are carried out in almost identical hyperparameter configuration except that learning rates may vary depending on the task. Learning rates for \pl~is generally large. Additionally, \pl~is more sensitive to learning rate. For example, HellaSwag performance vary for different.

\begin{table}[htbp]
    \centering\begin{tabular}{lllc}
        \hline
        \textbf{Experiment}        & \textbf{Method}         & \textbf{Hyperparameters}        & \textbf{Values} \\\hline
                                   & \multirow{5}{*}{Shared} & Optimizer                       & AdamW           \\
                                   &                         & Weight Decay                    & 0               \\
                                   &                         & Warmup Ratio                    & 0.1             \\
                                   &                         & LR Scheduler                    & Linear          \\
                                   &                         & Batch Size $\times$ Num$_{GPU}$ & 256             \\
        \hline
        \multirow{2}{*}{BoolQ}     & \lora                   & Learning Rate                   & 5e-4            \\
                                   & \pl                     & Learning Rate                   & \{5e-3,1e-3\}   \\\hline
        \multirow{2}{*}{MMLU}      & \lora                   & Learning Rate                   & 5e-4            \\
                                   & \pl                     & Learning Rate                   & \{5e-3,1e-3\}   \\\hline
        \multirow{2}{*}{HellaSwag} & \lora                   & Learning Rate                   & 5e-4            \\
                                   & \pl                     & Learning Rate                   & \{5e-3,1e-3\}   \\\hline
        \multirow{2}{*}{PIQA}      & \lora                   & Learning Rate                   & 5e-4            \\
                                   & \pl                     & Learning Rate                   & \{5e-3,1e-3\}   \\\hline
        \multirow{2}{*}{GSM8K}     & \lora                   & Learning Rate                   & 5e-4            \\
                                   & \pl                     & Learning Rate                   & \{5e-3,1e-3\}   \\\hline
    \end{tabular}
    \caption{Detailed hyperparameters used in our experiments. We vary learning rates depending on the dataset no matter the model scale. \pl~generally use larger learning rates. Gradient accumulation step is set to ensure the equality of total train steps for different models.}
    \label{tab-app-hparam}
\end{table}

\subsection{Prompt Templates}

We follow previous works to build prompts for tuning. More specifically, we follow QLoRA\mcite{qlora} for MMLU, MeZO\mcite{mezo} for BoolQ and LM-Eval Harness\mcite{eval-harness} for the rest. Despite improvement on benchmark scores in most of our experiments, the following prompts do not guarantee improvement on every model.

\begin{table}[htbp]
    \centering
    \begin{tabular}{lll}
        \hline
        Dataset                    & Task Type                        & Prompt                                     \& Target \\\hline
        \multirow{6}{*}{MMLU}      & \multirow{6}{*}{Multiple Choice} & <question>                                           \\
                                   &                                  & A.<samples>[0]                                       \\
                                   &                                  & B.<samples>[1]                                       \\
                                   &                                  & C.<samples>[2]                                       \\
                                   &                                  & D.<samples>[3]                                       \\
                                   &                                  & \em\color{blue}{A/B/C/D}                             \\\hline
        \multirow{2}{*}{BoolQ}     & \multirow{2}{*}{Classification}  & <passage><question>?                                 \\
                                   &                                  & \em\color{blue}{Yes/No}                              \\\hline
        \multirow{2}{*}{GSM8K}     & \multirow{2}{*}{Generation}      & Question: <question>                                 \\
                                   &                                  & Answer: \em\color{blue}{<answer>}                    \\\hline
        \multirow{6}{*}{HellaSwag} & \multirow{6}{*}{Multiple Choice} & <activity\_label>: <ctx> \_\_\_\_                    \\
                                   &                                  & A.<endings>[0]                                       \\
                                   &                                  & B.<endings>[1]                                       \\
                                   &                                  & C.<endings>[2]                                       \\
                                   &                                  & D.<endings>[3]                                       \\
                                   &                                  & Answer:\em\color{blue}{A/B/C/D. <endings>}           \\\hline
    \end{tabular}
    \caption{Prompt templates used for tuning. Prompts are adapted from QLoRA\mcite{qlora}, MeZO\mcite{mezo} and LM-Eval Harness\mcite{eval-harness}.
        <question> refers to fields of datasets  and {\em\color{blue}{{A}}} denotes target. During training, only loss on target part is calculated for back-propagation.
        For evaluation, we  use standard prompts provided in LM-Eval Harness and make no modification.}
    \label{tab-app-prompt}
\end{table}

\section{Throughput Measurement}

\subsection{Inference Throughput Measurement}

\subsubsection{Quantized Models On Mobile Phones}

We tested generation speed on mobile phones with MLC Chat\mcite{mlcllm}, currently only solution to running LLMs on mobile devices.
We use \llamat{7} (3bit) and RedPajama-3B (4bit) provided in the APP by default for benchmark.
We sample on each device for 20 times with inputs of 1024 tokens and report the  generation speed from built-in benchmarker. We use long sequences so that prefill stage is guaranteed to be FLOPS bound.

\begin{table}[htbp]
    \centering
    \begin{tabular}{l|cc|cc}
        \hline
        \multirow{2}{*}{Device Model} & \multicolumn{2}{c|}{\llamat{7} 3bit} & \multicolumn{2}{c|}{RedPajama-3B 4bit}                    \\\cline{2-5}
                                      & Prefill                              & Decode                                 & Prefill & Decode \\\hline
        iPhone 13 Pro Max             & 54.3                                 & 5.7                                    & 81.8    & 14.3   \\\hline
        Xiaomi 13                     & 58.9                                 & 8.1                                    & 88.1    & 16.3   \\\hline
    \end{tabular}
    \caption{Token generation speed tested on mobile phones with MLC Chat. Both prefill speed and decoding speed are reported with input sequences of 1024 tokens. Decoding speed is around 10\% of prefill speed due to memory boundness caused by non-batched computation.}
    \label{fig-app-mobilespeed}
\end{table}
Table \ref{fig-app-mobilespeed} lists prefill and decoding speed of . Decoding speed is much slower than prefill speed because the computation is not batched, thus memory bound.

\subsubsection{BF16 Models on a Single A100 80G GPU}

\begin{figure}[htbp]
    \centering

    \begin{subfigure}[b]{.48\textwidth}
        \includegraphics[width=\textwidth]{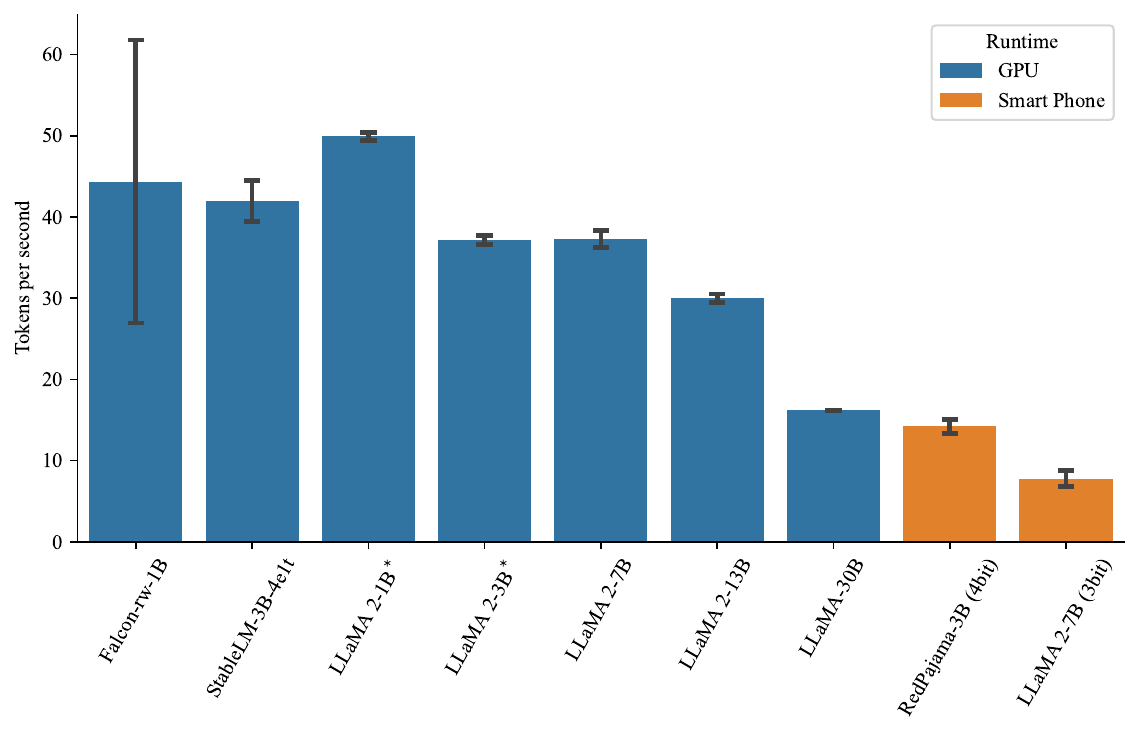}
        \caption{Token generation speed of decoding stage.}
    \end{subfigure}
    \begin{subfigure}[b]{.48\textwidth}
        \includegraphics[width=\textwidth]{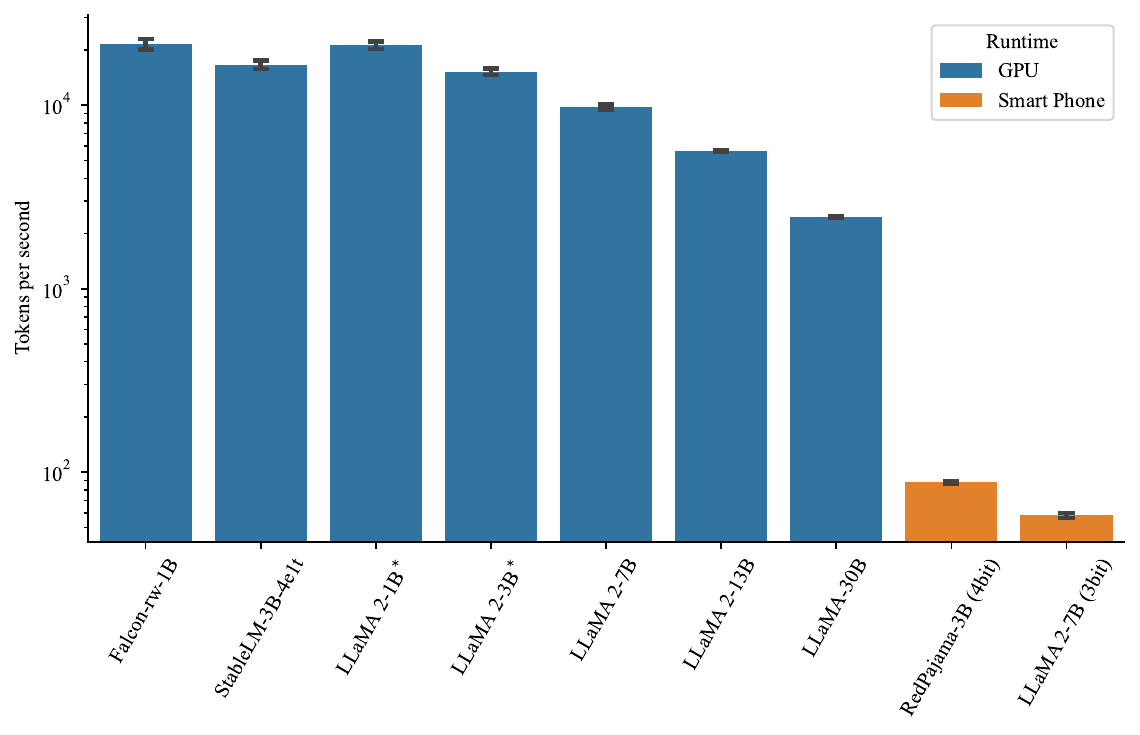}
        \caption{Throughput rate of prefill stage.}
    \end{subfigure}
    \caption{Decoding speed of models of scale between 1B and 30B parameters tested on $1 \times $A100 80G GPU. Models with $*$ are not official releases and are created by scaling transformer dimensions.}
    \label{fig-app-inferspeed}
\end{figure}

We use Huggingface Transformers\mcite{transformers} implementations to benchmark generation speed.
As for benchmark protocol, we follow vLLM\mcite{vllm} to sample elapsed time of prefilling 1024-token input and decoding 1024 new tokens.
Due to implementation, models can produce very results of large variance, \textit{e.g.} \falcono.
To rule out implementation discrepancy, we scale \llama~to 1B and 3B according to dimensions of \falcono~and \stablelm.
Obtained statistics are plotted on Figure \ref{fig-app-inferspeed}.
In decoding phase, 4bit quantized 3B model on smart phone is as fast as 30B model on a single A100 80 GPU.
For prefill stage, the gap between GPU and smart phones are significantly wider. Throughput of 30B model on GPU is almost 3000\% of quantized 3B model on smart phones.
The low throughput of prefill on smart phones hinders training on personal data.


\begin{table}[htbp]
    \centering
    \begin{tabular}{l|cc}
        \hline
        Model       & Batch Size & TPS    \\\hline
        \llamat{7}  & 16         & 1088.6 \\\hline
        \llamat{13} & 8          & 652.8  \\\hline
        \llamaa{30} & 2          & 296.0  \\\hline
    \end{tabular}
    \caption{Training throughput of llama series on a single A100 80G GPU.}
    \label{tab-app-tpsm-train}
\end{table}

\section{Transmission Budget Allocation of \lora$^S$}
\label{sec-app-loras}

\lora$^S$ denotes series of \lora~configurations that allows \lora~to reach predefined inference speed under the distributed architecture of \pl.
Used as baselines in our experiments, we set \lora$^S$ to reach 70\% of Cloud speed and derive the transmission budget per token $t$ for \lora.
We restrict adaptation modules to query, key and value for the same reason as \pl.
Therefore, the only variable to transmission amount is the number of adapted layers.
For \llamat{7}, to reach 70\% of Cloud speed, only 2 layers can be adapted.
For \llamat{13} and \llamaa{30}, the number of layers are 2, 3, respectively.
Therefore, we obtain the following potential configurations.
We test obtained configurations by tuning \llamat{7} on MMLU. Ranks of \lora~are set to 32 and $\alpha$ set to 32. Learning rates are 5e-4 and identical for all experiments.
\begin{table}[htbp]
    \centering
    \begin{tabular}{llll|c}
        \hline
        Target TPS            & \% Cloud TPS                 & Modules   & Layers & MMLU          \\\hline
        \multirow{5}{*}{25.6} & \multirow{5}{*}{$\sim $70\%} & q,k,v     & 0:2    & 47.4          \\
                              &                              & q,k,v     & 30:32  & 45.3*         \\
                              &                              & q,k,v     & 15, 31 & \textbf{50.4} \\
                              &                              & q,k,v     & 0, 15  & 48.7          \\
                              &                              & q,k,v     & 0, 31  & 45.9          \\\hline
        \multirow{3}{*}{11.7} & \multirow{3}{*}{$\sim $30\%} & q,k,v     & 0:16   & 55.5          \\
                              &                              & q,k,v     & 16:32  & 51.3          \\
                              &                              & q,k,v     & 0:32:2 & \textbf{55.8} \\\hline
        -                     & -                            & No-tuning &        & 45.3          \\\hline
    \end{tabular}
    \caption{MMLU scores of \llamat{7} tuned with different \lora$^S$ configurations. Notations of layers follow python list slicing. 0:16 in layers represents the first 16 decoder layers are adapted. Starred (*) scores indicate worse benchmark performance after tuning.}
    \label{tab-app-loras}
\end{table}

Table \ref{tab-app-loras} list MMLU scores of various \lora$^S$ configurations.
With a target TPS of 25, \lora$^S$ provides terrible tuning performances.
Adaptation on the 31\textit{st},32\textit{nd} layers and 1\textit{st},32\textit{nd} yields worse benchmark performances than no-tuning.
Increasing the number of adapted layers also increase the performances.
Adapting only the first 16 layers yield similar tuning performance to \lora$^P$.
Alongside performances of 30\% of Cloud TPS, adapting the first few layers gives better tuning performances.
Therefore, the general adaptation budget allocation strategy is to adapt shallow layers and final \lora$^S$ configurations are listed in Table \ref{tab-app-final-loras}

\begin{table}[htbp]
    \centering
    \begin{tabular}{l|cc|cc}\hline
        Model       & Modules & Layers & TPS  & \% Cloud TPS \\\hline
        \llamat{7}  & q,k,v   & 15,31  & 25.6 & 68.9         \\\hline
        \llamat{13} & q,k,v   & 19,39  & 19.5 & 70.0         \\\hline
        \llamaa{30} & q,k,v   & 29,59  & 12.4 & 74.6         \\\hline
    \end{tabular}
    \caption{Detailed adaptation configurations of \lora$^S$ for \llamat{7}, \llamat{13} and \llamaa{30} used in Section \ref{sec-exp}. Adapted modules are restricted to query,key and value to reduce transmission. The adapted layers are chosen with the insight that adapting shallow layers yield more performance gain. }
    \label{tab-app-final-loras}
\end{table}

\section{\pl~Throughput Estimation}
\label{sec-app-esti-tps}

\subsection{Inference Throughput}
\label{sec-app-infer-tps}

\begin{figure}[htbp]
    \centering

    \begin{subfigure}[b]{.48\textwidth}
        \includegraphics[width=\textwidth]{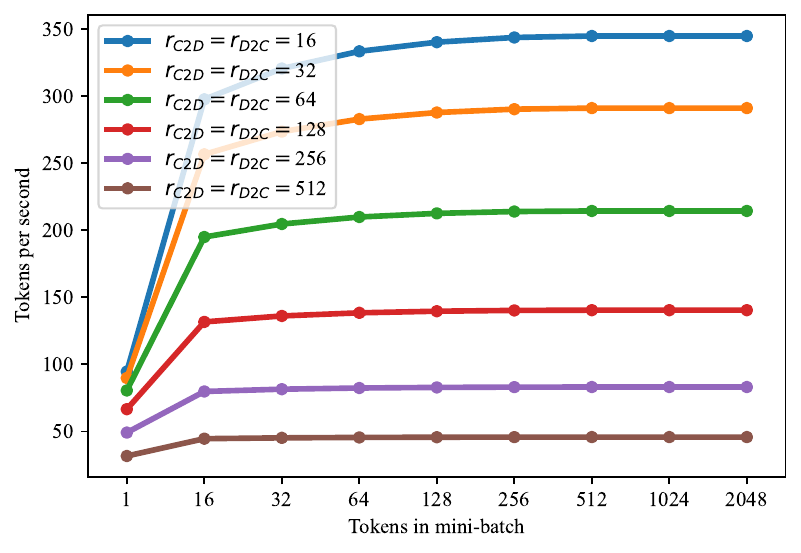}
        \caption{\llamat{13} Prefill.}
    \end{subfigure}
    \begin{subfigure}[b]{.48\textwidth}
        \includegraphics[width=\textwidth]{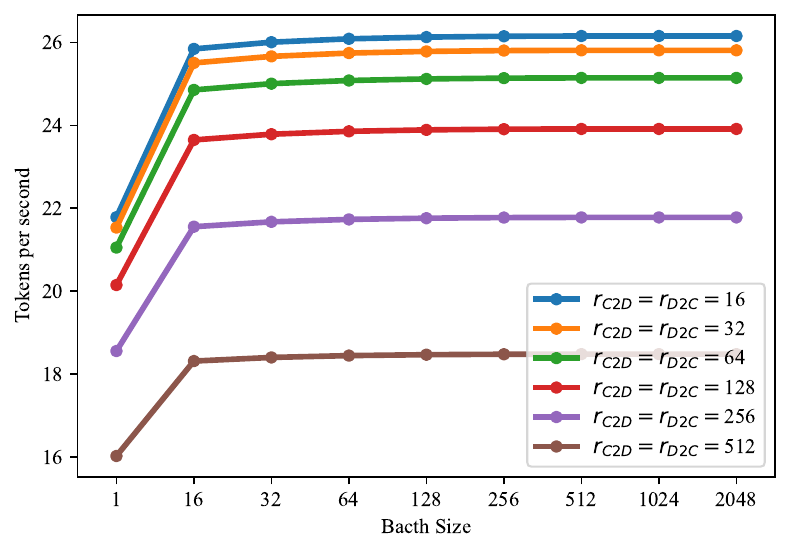}
        \caption{\llamat{13} Decoding.}
    \end{subfigure}
    \begin{subfigure}[b]{.48\textwidth}
        \includegraphics[width=\textwidth]{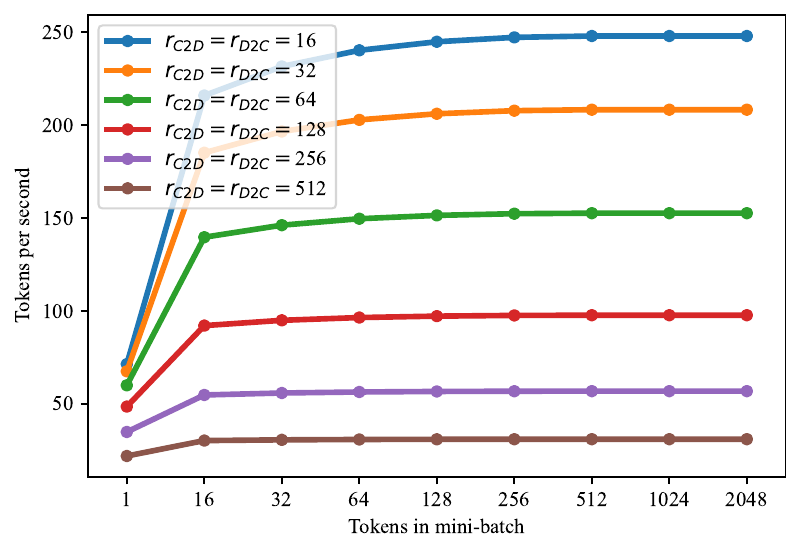}
        \caption{\llamat{30} Prefill.}
    \end{subfigure}
    \begin{subfigure}[b]{.48\textwidth}
        \includegraphics[width=\textwidth]{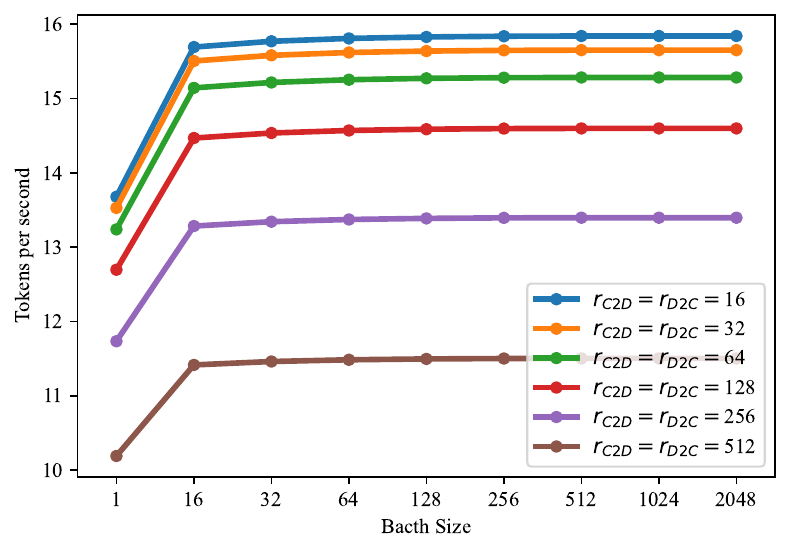}
        \caption{\llamat{30} Decoding.}
    \end{subfigure}
    \caption{Estimated inference throughput of \pl~on different base models. }
    \label{fig-app-infer-throughput}
\end{figure}

\subsection{Training Throughput}

\begin{figure}[htbp]
    \centering

    \begin{subfigure}[b]{.48\textwidth}
        \includegraphics[width=\textwidth]{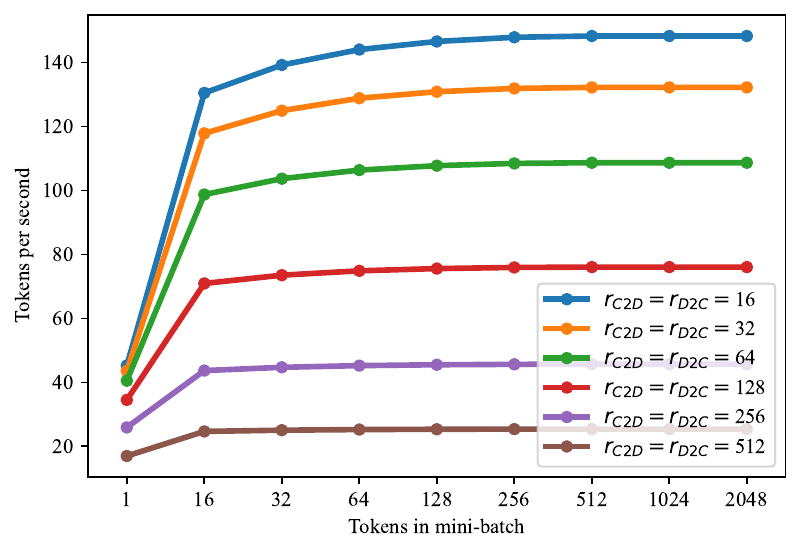}
        \caption{Training throughput (\llamat{13}).}
    \end{subfigure}
    \begin{subfigure}[b]{.48\textwidth}
        \includegraphics[width=\textwidth]{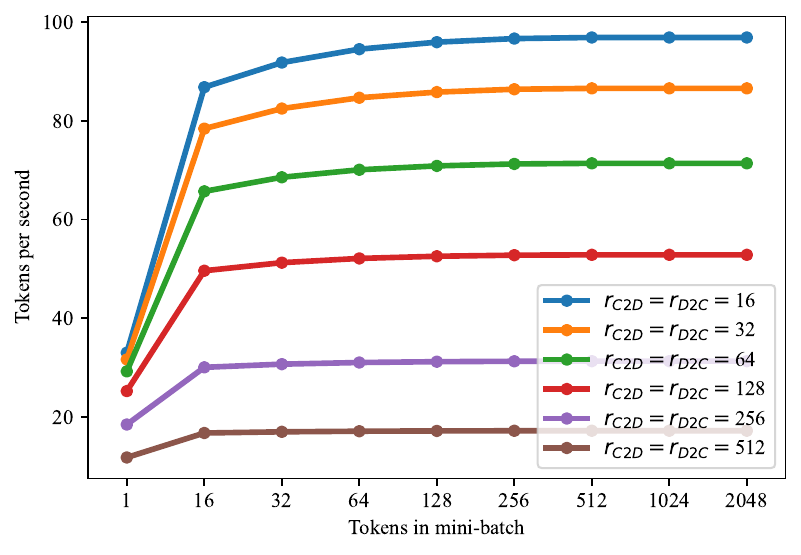}
        \caption{Training throughput (\llamaa{30}).}
    \end{subfigure}
    \caption{Estimated training throughput of \pl~on different base models. }
    \label{fig-app-train-throughput}
\end{figure}
\end{document}